\documentclass[runningheads]{llncs}
\usepackage{graphicx}
\usepackage{times}  
\usepackage{helvet}  
\usepackage{courier}  
\usepackage{url}  
\usepackage{algorithm}
\usepackage{graphicx}
\usepackage{varwidth}
\usepackage[noend]{algpseudocode}

\usepackage{times}
\usepackage{listings}
\usepackage{caption}

\usepackage{amsmath}
\begin{document}
\title{Action Model Acquisition using LSTM}
%
%
\author{Ankuj Arora \and
Humbert Fiorino \and
Damien Pellier\and
Sylvie Pesty}
\authorrunning{Arora et al.}
%
\institute{Univ. Grenoble Alpes, CNRS, Grenoble INP, LIG, 38000 Grenoble, France
\email{firstname.lastname@univ-grenoble-alpes.fr}\\
\url{https://www.liglab.fr}}
\maketitle              

\begin{abstract}
In the field of Automated Planning and Scheduling (APS), intelligent agents by virtue require an action model (blueprints of actions whose interleaved executions effectuates transitions of the system state) in order to plan and solve real world problems. It is, however, becoming increasingly cumbersome to codify this model, and is more efficient to learn it from observed plan execution sequences (training data). While the underlying objective is to subsequently plan from this learnt model, most approaches fall short as anything less than a flawless reconstruction of the underlying model renders it unusable in certain domains. This work presents a novel approach using long short-term memory (LSTM) techniques for the acquisition of the underlying action model. We use the sequence labelling capabilities of LSTMs to isolate from an exhaustive model set a model identical to the one responsible for producing the training data. This isolation capability renders our approach as an effective one.

\end{abstract}


\section{Introduction}

Automated Planning and Scheduling (APS) is anchored on the formulation of a plan: a sequential and orchestrated execution of actions (represented as action signatures) which effectuates state transitions in the domain, gradually propelling it towards a predetermined goal. These action signatures, alongwith their preconditions and effects, constitute a ``script'', also known as the \textit{action model}, which governs their execution, and consequently: plan formation.  As planning domains become larger, the challenges surrounding the conceptualization of these action models becomes more magnified.
In such domains, planners now leverage advancements in machine learning (ML) techniques to learn action models from training data i.e. learn action preconditions and effects from their signatures.

In simpler domains, the possibility of learning a rugged preliminary action model which can serve as the starting point for domain experts to complete and enrich serves great benefits \cite{DBLP:conf/aips/GregoryL16}. However, in more complex domains, such as the case of dialogue planning for Human Robot Interaction (HRI), this approach may fall short \cite{arora2016review}. The interleaving between various body gestures and utterances which constitute multimodal dialogues is highly delicate. In such cases, since the intricacies and subtleties are so barely identifiable, even expert intervention would not be sufficient to render the model reusable. Moreover, this partial model generation points to an incapacity to subsequently plan without expert intervention. This intervention in the case of multimodal HRI is insufficient, thus demonstrating a strong need for a very accurately learnt model in order to subsequently plan.
Our goal is to output a high quality model structurally identical to the hand woven model using minimum domain knowledge in the input. Our approach, using only state-action interleaved traces (alternating state and action representations) in the input, tackles the issue of incomplete model generation as follows: beginning with an exhaustive generation of all possible candidate action models, we gradually narrow down to the speculated ideal model (empirically isolated model speculated to be identical to the hand woven model) by exploiting (i) the inter-action dependencies in traces using pattern mining techniques and (ii) the sequence labeling capabilities of recurrent neural networks.
Our approach is called PDeepLearn, which stands for \textbf{P}DDL domain \textbf{Deep} \textbf{Learn}er. As the name suggests, the approach learns PDDL  \cite{mcdermott1998pddl} domain models with the help of deep learning techniques. In particular, the noteworthy successes of LSTM \cite{hochreiter1997long}, a deep learning technique being increasingly used in learning long range dependencies in the fields of speech and handwriting recognition \cite{lecun2015deep}, are exploited. A plan is an orchestrated execution of a series of actions which are by virtue interdependent in order to execute. Since a plan exhibits long-term dependencies, and the LSTM by virtue exploits the same; there is a direct link between the principle of operation of the LSTM and plan execution, which this paper seeks to explore. This paper is divided into the following sections: we present some related work in section~\ref{relatedW}, followed by the definition of our learning problem in section~\ref{preliminaries}. We then detail the functioning of the PDeepLearn system in section~\ref{approach}, and present our empirical evaluations over four artificial domains in section~\ref{evaluation}. We conclude the paper with some perspectives and future work in section~\ref{conclusion}.

\section{Related Work}
\label{relatedW}
 Our goal is, with the least amount of domain knowledge possible, to output a high quality model structurally identical to the hand woven model used to generate the training data. This is in contrast with several approaches in the literature \cite{DBLP:conf/icaart/McCluskeyCRW09,mccluskey2002interactive} which require a substantial amount of user or expert input. Some works that have demonstrated relatively high proximity of the learnt model to the hand woven one \cite{DBLP:conf/aips/CresswellG11,DBLP:conf/ijcai/ZhuoK13} do not reach to the extent of structural identicalness to the hand woven model. Other approaches in the literature that subsequently plan with learnt operators \cite{DBLP:conf/aaai/JimenezFB08,DBLP:conf/aips/GregoryL16,DBLP:conf/ijcai/ZhuoNK13} manage to plan a fraction of the test problems, thus highlighting the distance of the learnt model from the hand-woven one. In other words, these models are short of vital information, and are thus incomplete. We also draw comparisons with approaches that exploit pattern mining techniques. 

Various approaches in the past have exploited
patterns and inter-action dependencies in traces to learn action models. This includes treating operator sequences in the form of n-grams \cite{muise2009exploiting} or macro-actions \cite{DBLP:conf/ijcai/ZhuoNK13}.
 Pattern mining techniques have also been used to find interesting correlations between actions that figure in the traces. For example, the ARMS \cite{yang2007learning} system finds frequent action pairs that share a common set of parameters from plans using the Apriori algorithm \cite{agrawal1994fast}. However, none of these approaches learn a model in high syntactic proximity with the handwoven action model.
Artificial neural networks in their most basic form (i.e. the perceptron) have previously been used in the literature \cite{DBLP:conf/uai/MouraoZPS12,mourao2008using,DBLP:conf/ecai/MouraoPS10} to learn action models. We are, however, the first ones in our knowledge to employ deep learning techniques to learn action models.

\section{Preliminaries and Problem Formulation}
\label{preliminaries}
In the field of APS, agents interact with the environment by executing actions which change the state of the environment, gradually propelling it from its initial state towards the agents' desired goal. In the classical representation, both the world state and actions are pre-engineered and constituted of properties called \textit{predicates}. \textit{States} are defined to be sets of ground (positive) predicates. Here, each action a$\in$A where A = $\left\lbrace a_{1}, a_{2}, \ldots a_{n} \right\rbrace$, $n$ being the maximum number of actions in the domain. We use actions and operators interchangeably in our context. These actions constitute a corpus which serves as a blueprint for these actions, called the \textit{action model}. 
An \textit{action model} $m$ is the blueprint of all the domain-applicable actions belonging to the set $A$. Each action is defined as an aggregation of : (i) the action name (with zero or more typed variables as parameters), and (ii) three lists, namely ($pre, add$ and $del$). These are: the \textit{pre} list (predicates whose satisfiability determines the applicability of the action), \textit{add} list (predicates added to the current system state by the action execution) and the \textit{del} list (predicates deleted from the current system state upon action execution), respectively. A \textit{planning problem} is a triplet P = $(s_0, g, m)$ composed of (i) the initial state of the world $s_o$, (ii) the desired goal $g$ and (iii) the action model $m$. All the aforementioned elements in this section contribute to the formulation of a \textit{plan}. A \textit{plan}, given $P$, is an interleaved sequence of actions and states represented as: $\pi=[s_0, a_1, s_1, a_2,\ldots, a_n, g]$ that drives the system from the initial state to the goal. Each alternating state-action sequence, complete with initial state and goal information, constitutes a \textit{trace}. An aggregation of these traces constitutes a trace set \textit{T}. Any sequence of actions is likely to exhibit implicit patterns and regularities within them. These patterns may be of the form of frequently co-occurring actions, which indicate the possible presence of inter-action dependencies and relationships. These relationships can be uncovered by means of pattern mining techniques to facilitate the process of learning. The aforementioned trace set \textit{T} can thus be treated as a \textit{Sequence Database} (SD) in the pattern mining domain. An SD is a set of sequences $T=[s_1, s_2,\ldots, s_s]$ where each sequence represents a trace. Each sequence is an ordered list of items (in our context, each item in the sequence is equivalent to an action) $s_i=[a_1, a_2,\ldots, a_n]$ where $(a_1, a_2,\ldots, a_n) \in A$. A sequential rule $a_x\to a_y$ is a relationship between two actions $a_x$, $ a_y \in A$ such that
if $a_x$ occurs in a sequence, then $a_y$ will occur successively in the same sequence. Two measures are defined for sequential rules. The first one is the \textit{sequential support}. For a rule $a_x\to a_y$, it is defined as $sup(a_x\to a_y) = |a_x\to a_y| / |T|$. The second one is the \textit{sequential confidence} and it is defined as $conf(a_x\to a_y) = |a_x\to a_y| / |a_x|$, where $|a_x\to a_y|$ represents the number of times the $a_x$ and $a_y$ appear in succession in $T$, and $|a_x|$ represents the number of times $a_x$ appears in $T$. The values of the confidence and support of $a_x$ and $a_y$ is directly proportional to the frequency of the sequential co-occurrence of $a_x$ and $a_y$ in the SD \cite{fournier2012mining}.

Given the aforementioned information, the problem of learning action models can be formulated as follows: given a set of observed plan traces \textit{T}, each plan trace consisting of (i) the initial state, (ii) state-action interleavings and (iii) the goal; the objective is to learn the underlying action model $m$ which best explains the observed plan traces. APS defines a number of synthetic simplified planning domains, out of which we use the \textit{gripper} domain to exemplify our approach. In this domain, the task of the robot is to move an object from one room to the other. The principal actions in this domain (mentioned on the left of the image) are \textit{(move, pick, drop)}. The principal predicates include: \textit{at} (true if object \textit{?obj} is present in room \textit{?room}), \textit{at-robby} (true if robot \textit{?r} is present in room \textit{?room}), \textit{free} (true if gripper \textit{?g} of robot \textit{?r} is free), \textit{carry} (true if object \textit{?obj} is carried in gripper \textit{?g} of robot \textit{?r}).

\section{The PDeepLearn Approach}
\label{approach}
\begin{figure}[h]
  \centering
  \includegraphics[width=0.7\linewidth]{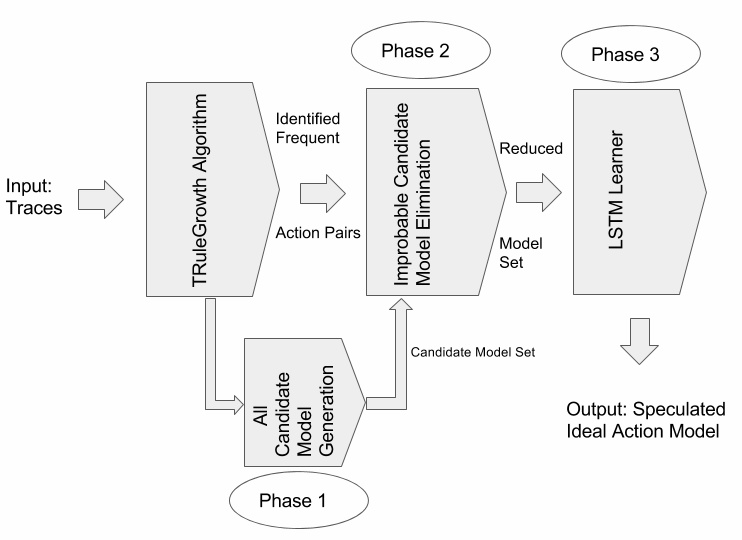}
  \captionof{figure}{PDeepLearn approach phases}
  \label{algo}
\end{figure}

The approach followed in this paper is divided into three phases, which are summarized in figure~\ref{algo}. In the first phase, we enumerate all possible candidate action models. In phase two, we identify frequent action pairs using pattern mining algorithms. \textit{Action pair constraints} are applied to the frequent action pairs to eliminate improbable candidate models. In the final phase, LSTM based techniques are used to label action sequences with the intent of identifying the speculated ideal model which produces the best labelling accuracy.  
This model is proved to be structurally identical to the hand woven model used to produce the input traces serving as our training data.



\subsection{Candidate Model Generation}
\label{candgen}
In this subsection, all possible candidate action models are generated from the input plan traces. Firstly, each trace is taken one by one, and each action as well as each predicate in the initial, goal and intermediate states of the plan trace is scanned to substitute the instantiated parameters with their corresponding variable types. This builds the corresponding generalized action schema $A_s$ and predicate schema $P_s$ respectively. Each action in the action schema is associated with its relevant predicates from the predicate schema, where a predicate $p_{a_i}\in{P_s}$ is said to be \textit{relevant} to an action $a_i\in{A_s}$ if they share the same variable types. For example, the predicate \textit{at-robby (?robot, ?room)} is relevant to the action \textit{find (?robot, ?object, ?room, ?gripper)} as both of them contain the parameter types \textit{(robot, room)} in their signatures. The set of relevant predicates to an action $a_i$ can be denoted as $relPre_{a_i}$.
With schemas $A_s$ and $P_s$, a candidate action dictionary is built. Each \textit{key} in the dictionary is the name of the action ($a_i\in{A_s}$) and the \textit{value} is a list of all predicates relevant to that particular action ($relPre_{a_i}\in{P_s}$). All possible combinations of relevant predicates per key are represented as elements of a candidate action set $CAS$. We represent the elements of the candidate action set $CAS_{a_i}$ of each action $a_i\in A_s$ in the format $((pre), (add), (del))$. The cartesian product of all these candidate action sets per action constitute the set of all possible candidate models $M$ for the domain. The size of $M$ is directly proportional to the complexity of the domain i.e. number of actions and predicates constituting the domain. More precisely, for a domain containing $n$ actions, each with $m$ relevant predicates, the complexity of the generation procedure comes out to be $(\prod_{i=1}^n \binom{m}{i})^3n$. We use the \textit{gripper} domain to illustrate the aforementioned steps, taking the example of the \textit{pick} action. The candidate model generation procedure for the \textit{pick} action is represented in Listing~\ref{fig:generation}.\\

\begin{lstlisting}[language=C,frame=,caption=Candidate model generation procedure for the \textit{pick} action of the \textit{gripper} domain,label=fig:generation,captionpos=b]
|\textbf{trace snippet before generation phase::}|
[|\textbf{(action)}| (pick robot3 ball5 room1 rgripper3) |\\\textbf{(state)}| ((at ball5 room1) |(at-robby (...)))||(carry (...))|]
|\textbf{trace snippet after generation phase::}|
[|\textbf{(action)}| (pick robot ball room rgripper) |\\\textbf{(state)}| ((at ball room) (at-robby (...))) |(carry (...)|]
|\textbf{Relevant predicates for action}| |$\textit{pick}$|: 
(at, at-robby, carry)
|\textbf{Candidate action set for action}| |$\textit{pick}$|:
{((at), (carry), (|$\neg$|at-robby)), ((at-robby), (carry), (|$\neg$|at)),...}
\end{lstlisting}



In order to qualify as legitimate, each model $m\in M$ must satisfy some \textit{semantic constraints} proposed by STRIPS \cite{fikes1971strips}. As per these constraints, for each action $a\in{A_s}$ and predicate $p\in{P_s}$ in the model \textit{m}: (i) $p$ cannot be in the \textit{add} list and the \textit{del} list at the same time, and (ii) \textit{p} cannot be in the \textit{add} list and the \textit{pre} list at the same time. Models which do not adhere to these constraints are deleted from the set $M$. This step is quintessential in performing a first round of filtration, thus weeding out a large number of improbable models and leaving us with more accurate ones which are taken to the next step of data mining.

\subsection{Sequence Pattern Mining}

We hypothesize that if a sequential pair of actions appears frequently in the traces, there must be a reason for their frequent co-existence. We are thus interested in the branch of pattern mining algorithms which treat frequent sequential action pairs. Other algorithms in the pattern mining literature, like the Apriori algorithm \cite{agrawal1994fast} (used in ARMS \cite{yang2007learning}) are not equipped to satisfy this requirement. Bearing this requirement in mind, the input traces are parsed and fed to an algorithm called TRuleGrowth \cite{fournier2012mining}, an algorithm used for mining sequential rules common to several sequences that appear in a sequence database. It belongs to a family of algorithms constituting a data mining library called SPMF \cite{fournier2014spmf}, a comprehensive offering of implementations of data mining algorithms.
The inputs to TRuleGrowth are (1) a sequence database $T$, (2) \textit{support} (a value in [0, 1]) and (3) \textit{confidence} (a value in [0, 1]). Given a sequence database $T$, and the parameters $support$ and $confidence$, TRuleGrowth outputs all sequential rules having a support and confidence higher than $support$ and $confidence$ respectively. Starting with 10 traces, we consistently double the number of traces till we reach 700. In the process, we identify frequent rules (action pairs) which consistently maintain the values of \textit{confidence} and \textit{support} over an increasing number of traces.

\subsection{Candidate Model Elimination}\label{cElemination}
After identifying frequent action pairs in the previous step, this step filters elements of the candidate action set for the actions constituting the frequent action pairs. These frequent pairs are suspected to share a ``semantic" relationship among themselves, in terms of a correlation between their $pre$, $add$ and $del$ lists. These relationships serve as \textit{action pair constraints}, and have been proposed by the ARMS \cite{yang2007learning} system, serving as heuristics to explain the frequent co-existence of these actions. These heuristics produce good results in the case of the ARMS system, which serves as incentive for re-using the same. More precisely, if there is an action pair $(a_i, a_j), 0 \leq i < j \leq (n-1)$ where $n$ is the total number of actions in the plan; and $pre_i$, $add_i$ and $del_i$ represent $a_i$'s \textit{pre}, \textit{add} and \textit{del} list, respectively: (i) a predicate $p$ which serves as a precondition in the \textit{pre} lists of  both $a_i$ and $a_j$ cannot be deleted by the first action, (ii) a predicate $p$ that is added by the first action $a_i$ appears as a prerequsite for the second action $a_j$ in an action pair, and (iii) a predicate $p$ that is deleted by the first action $a_i$ is added by $a_j$.

We use these constraints in a different fashion than in the original ARMS implementation. All pairs of elements in the candidate action set which satisfy one or more of the aforementioned constraints will be retained for the following learning step. For example, if \textit{(move,pick)} is identified to be a frequent action pair and $((at-robby), (), (\neg at-robby)) \in CAS_{move}$; and $((carry), (at-robby), (\neg at)) \in CAS_{pick}$ are two elements; then these two elements are retained in their respective action sets as they satisfy the third action pair constraint. Thanks to this heuristic, all actions which occur in frequent pairs will have pruned candidate action sets for the learning phase; thus speeding up learning by narrowing down the initial candidate set. For each action pair $(a_i, a_j)$ with $|CAS_{a_i}|=m$ and $|CAS_{a_j}|=n$, the complexity of this step is $(m\times n)$. Please note that all the actions which not figure in the frequent sequential pairs will not undergo this pruning phase, and will directly pass to the following learning phase.

\subsection{LSTM based Action Classification for Model Learning}
Each action in a plan execution sequence is influenced by previously executed actions, and will influence future actions further down the sequence. Thus, extracting patterns from sequences of previously executed actions is likely to provide strong evidence to predict the label of the next action in the chain. Through the medium of sequence labelling, we can narrow down and isolate the speculated ideal model among the reduced candidate set, as it would hypothetically be the one delivering the highest prediction rate amongst all models in the set. These dependency-driven and chained action executions inspire our investigation of long short-term memory networks (LSTMs) \cite{hochreiter1997long} for action sequence labelling, with the intent of identifying the speculated ideal model.
In the following subsections, we present our data encoding method for the input and output vector of the LSTM, and an overview of the training and validation phases of our framework.

\subsubsection{Data Encoding for Labelling of Action Sequences}
We use the sequence labelling capabilities of LSTM to identify the speculated ideal model from the reduced candidate set $M$. The sequence labelling in this case is in fact a classification of the most likely action that succeeds a given one. The input to our LSTM system is a large corpus comprising vector representations of each action of each trace. Each trace is taken one by one, and the comprising actions are sequentially encoded into input and output vectors; thus producing a large corpus of vectors. Each  action of each trace is represented by two distinct vectors: an input vector which encodes the action, and an output vector which classifies the next action succeeding the action currently being encoded (we refer to this action as the current action for the sake of simplicity). These vectors serve as the input and output respectively to the LSTM cells, the encoding of which represents the heart of this section. This corpus of vector representations is divided into a training and validation set aimed at training the LSTM on the training set and gauging its performance on the validation set. At the output of this learning system, we obtain an accuracy of prediction on the folds of validation data. The encoding of the input and output vectors is represented in the following paragraphs.

\begin{figure}[h]	
\centering
\includegraphics[width=4.0in]{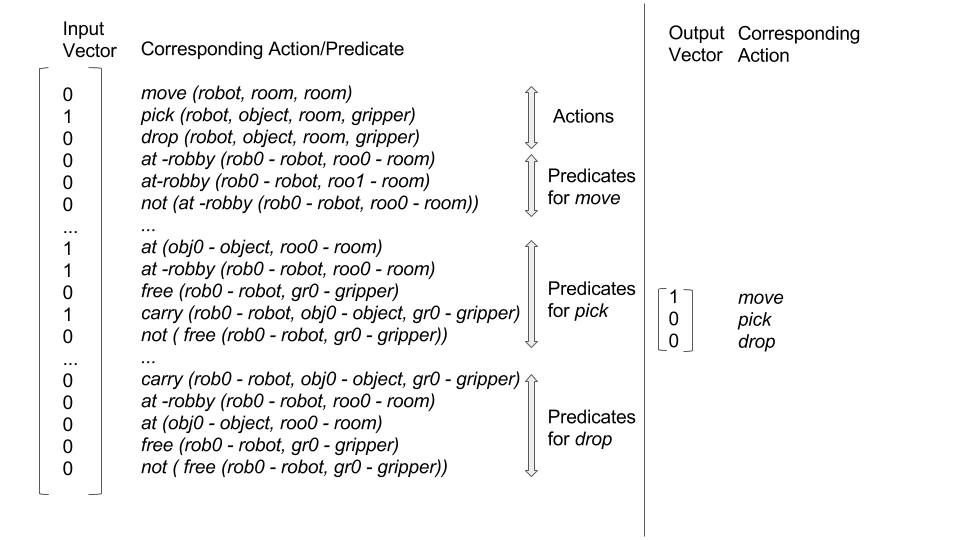}
\caption{Input and output vectors to the learning system for in training phase for currently encoded action \textit{pick} and successive action \textit{move}.}
\label{fig:sub1}
\end{figure}

The input vector representing an action in a trace is encoded in the following fashion. It is divided into two sections: one section which labels the entire set of actions in the domain, and the other which labels the relevant predicates for the actions in the domain. In the first section, there is a slot for each action in the domain. The slot for the action currently being encoded is labeled as 1, and the slots for the remaining actions in the domains are labeled as 0. Thus if ${(a_1,a_2,\ldots,a_n)}\in A$ is the set of domain-applicable actions, the first \textit{n} elements of the vector will be representing this first section, with the entry for the action currently being encoded being switched to 1, the other $(n-1)$ slots for the remaining $(n-1)$ actions being kept at 0. Once this first section has been assigned, we dedicate in the second section blocks of elements in the vector specific to each action in the domain. Thus for $n$ actions in the domain, there are $n$ different blocks (plus the one block for all the domain applicable actions as explained above). Each action-specific block contains one entry for each predicate relevant to that particular action. Revisiting the example in section~\ref{candgen} for the \textit{gripper} domain, if \textit{[at-robby (rob0 - robot, roo0 - room), at (obj0 - object, roo0 - room)]} are two predicates relevant to the action $pick$, they will constitute two entries in the $pick$ action block. We thus create action-specific blocks and for each action, assign entries for all the predicates relevant to that action, replicating this scheme for every action in the domain. Thus, the number of blocks for the input vector stands at $(n+1)$. The dimension $d$ of this input vector is directly proportional to the number of actions in a domain, as well as the number of predicates relevant to each action. The dimension $d$ of a vector for a specific domain will always remain the same, with the switching of a slot from 0 to 1 in the vector signalling the execution of a particular action. If $(a_1, a_2,\ldots, a_n) \epsilon {A_s}$ represents the action schema, the dimension of the input vector is given as:
\begin{equation} d = n+\sum_{i=1}^{n}relPre_{a_i}\end{equation}
Here $relPre_{a_i}$ are the number of relevant predicates for the action $a_i$.\\
The output vector predicts the label of the action that follows the action currently being encoded in the trace. Very much like the input, the output is encoded as a binary vector. It consists of a single block which has as many slots as the number of actions in the domain, one for each action. The slot representing the succeeding action to the action being currently encoded is set to 1, the others being set to 0. Revisiting the example in section~\ref{candgen} for the \textit{gripper} domain, the action currently being encoded is \textit{pick} and the next action in the trace is \textit{move}. The input and output vectors for the \textit{pick} action are represented in Figures~\ref{fig:sub1} and~\ref{fig:sub2}. While the number of actions in a trace, and thus the number of vectors representing all the actions in a trace may vary, the LSTM requires a fixed sized input. This is ensured by calculating the maximum trace length \textit{batchLen} (maximum number of actions per trace for all the traces) for all the traces, and padding the shorter lists with \textit{d}-dimensional vectors filled with zeros. This padding is done for all the traces till all the traces have the same \textit{batchLen} number of actions. The same padding procedure is adopted for the output vectors. 

\subsubsection{Training and Validation Phase}
\label{trainingVal}

\begin{figure}[h]	
\centering
\includegraphics[width=4.0in]{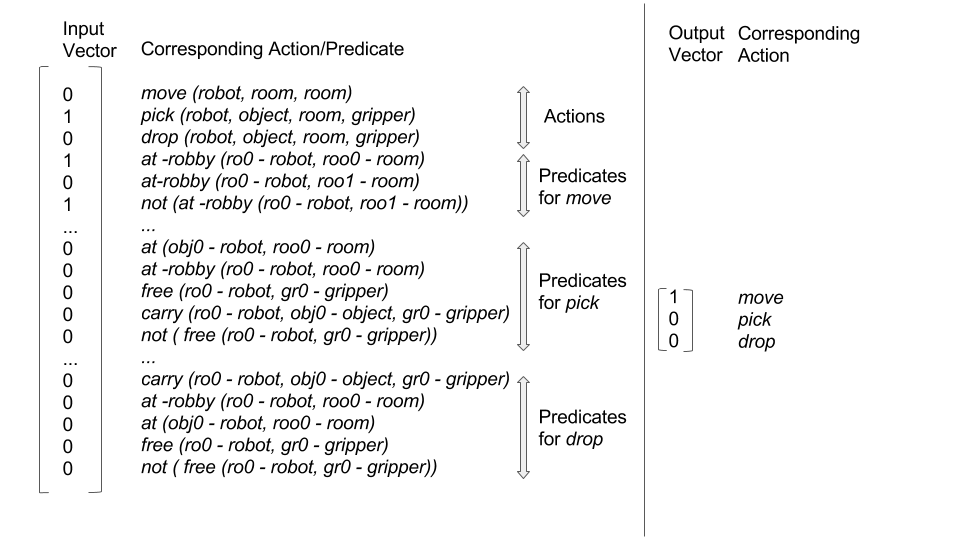}
\caption{Input and output vectors to the learning system for the validation phase with a candidate model for currently encoded action \textit{pick} and successive action \textit{move}}
\label{fig:sub2}
\end{figure}

The training and validation phases vary in the fashion that the slots of the input vector are labelled. This is because of the fact that the objective of the validation phase is to test the individual models to isolate the one with the highest labelling accuracy. In the training phase, two sections of the vector are filled out to represent the action currently being encoded. In the first section, the label of the action currently being encoded is set to 1, the rest being kept at 0. In the second section, slots of the predicates relevant to the currently encoded action which occur in the current state (that is, are potential preconditions for the action about to be executed or about to be encoded) is set to 1. Also, all slots the newly introduced predicates as a result of the execution of the current action (i.e. the difference between the current state and the next state) which are relevant to the current action are set to 1, the other slots being kept at 0. The other blocks are kept switched off at 0 as well. This mechanism of encoding the training vector is elaborated as follows. Revisiting the example from Listing~\ref{fig:generation}, if the action currently being encoded is \textit{pick}, the state prior to the application of the action has the following predicates: \textit{at, at-robby}. Both these predicates are relevant to the current action \textit{pick}. These predicates serve as potential precursors to the application of the action \textit{pick}, and are suspected to be the preconditions for the action. Thus, in the training vector, for the block specific to the action \textit{pick}, the slots representing these two predicates are set to 1. The execution of the action introduces new predicates into the state represented after the action execution, namely the predicate \textit{carry}. This predicate is relevant to the action \textit{pick}, and can be suspected to constitute the effects of this executed action. Thus, in the case of the training vector, the slot representing the \textit{carry} predicate of the \textit{pick} action is set to 1. In conclusion, all the predicates slots relevant to the action being executed which are suspected to constitute the preconditions and effects of the model are switched to 1, while the blocks of the other actions in the training vector are kept at zero. This vector is similarly constructed with every successive action being encoded.\\
In the validation phase, the objective is to test models sampled from the candidate set $M$ in order to zero down to the speculated ideal model. Promising models are sampled from $M$ by passing each model one by one though a planner and tested for their capability to solve a unitary problem.  Models that fail to find a solution are discarded. Models that find a solution are retained in $m'$. The encoding for each of these models for validation with the LSTM is done in the following fashion. The first section is represented in the same way, with the label of the current action set to 1. The sections are, however, labelled differently. In this case, the slots in the vector which correspond to the relevant predicates present in the current action of the current model being evaluated are set to 1. For example, as illustrated in the Figure~\ref{fig:sub2}, if one of the candidate models are represented by ($move$: (at-robby, $\neg$(at-robby)), $pick$: (free, carry), $drop$: (carry, $\neg$(not-free))), then the slots in the vector for the $move$ action which represent the predicates $(at, \neg (at-robby))$ are switched to 1, the rest of the predicates being kept at 0. This validation is done for each of the models of the set $m'$.





\begin{table}[]
\centering
\label{truletable}
\begin{tabular}{|l|l|l|l|l|}
\hline
Domain    &\begin{tabular}[c]{@{}l@{}} Initial Number\\of Candidate\\ Actions\end{tabular} &\begin{tabular}[c]{@{}l@{}} Final Number\\of Candidate\\ Actions post\\ Pruning\end{tabular} &\begin{tabular}[c]{@{}l@{}} Percentage\\ Reduction\\(\%)\end{tabular} \\ \hline
Satellite & 1633                                & 155                                 & 90.6                \\ \hline
Mprime    & 2834                                & 466                                 & 83.55               \\ \hline
Gripper   & 292                                 & 6                                   & 97.94               \\ \hline
Depots    & 2913364                             & 10240                               & 99.65              \\ \hline
\end{tabular}
\caption{Model Pruning Results for PDeepLearn}
\end{table}

\section{Evaluation}
The objective of the evaluation is to identify the speculated ideal model which can then be syntactically compared with the hand woven model for its identicalness. This section presents the evaluation results for the PDeepLearn system tested on four domains, namely: \textit{satellite, mprime, gripper} and \textit{depots}. The section is divided into three subsections which correspond to the three phases of the PDeepLearn algorithm.

\label{evaluation}
\subsection{All Candidate Model Generation}
The cumulative total of all candidate actions for each domain are represented in the second column of Table 1. The size is proportional to the complexity of the domain i.e. the number of predicates relevant to each action. Although the complexity of the generation phase is proportional to the domain complexity, the generation time for each of the domains, as shown in column 3 of Table 2, is fairly negligible.


\subsection {Sequence Pattern Mining}


The traces must be parsed and converted into a specific format which can serve as an input to the TRuleGrowth algorithm \cite{fournier2012mining}. The emerging rules with their confidence and support are recorded. The rules maintaining a consistently high value of confidence and support (empirically determined) with increasing number of traces for domains are recorded, since they represent a high-frequency sequential pair of actions, and are further pruned with \textit{action pair constraints} detailed in Section~\ref{cElemination}.

\subsection{Candidate Model Elimination}

In accordance to the \textit{action pair constraints} mentioned in Section~\ref{cElemination}, elements of the candidate action sets of the frequent actions which satisfy atleast one of the three constraints are retained, while the others are deleted. This elimination produces a reduced candidate set of models. The resulting number of candidate actions and the eventual reduction in the number of candidate actions of this phase are represented in the second and third column of the table 1. The highest reduction can be seen in the \textit{depots} domain, while the least reduction can be seen in the \textit{mprime} domain. This can be explained by a stronger correlation among actions of the \textit{depots} domain than among the actions of the \textit{mprime} domain. The reduced candidate set is then passed on to the LSTM.


\subsection{LSTM Based Speculated Ideal Model Identification}
This subsection is used to isolate the speculated ideal model from the sampled candidate model set $m'$ (sampling time recorded in Table 2). In this work, we explore two hyperparameters: the number of hidden units (set between {(100, 200))}, and the dropout rate \cite{lecun2015deep} (set between {(0.5, 0.75))}; both of which have significant potential to influence the proposed LSTM-based labelling predictions. Other than these, we use a softmax layer for classifying given sequences of actions. The batch size is set to $batchLen$. We also use categorical cross entropy for the loss function and an adam optimizer (gradient descent optimizer). Each dataset for each of the 4 evaluated domains consists of 700 examples each, which is divided using five-fold cross validation. Every training example is presented to the network 10 times i.e. the network is trained for 10 \textit{epochs}. The results are summarized in the Table 3 and are the obtained with the speculated ideal model (700 traces, 128 hidden units, 0.8 dropout rate), which produces the highest accuracy as compared to other models in the sampled set. The accuracy represented here is the \textit{validation accuracy} (accuracy measured on the validation set), which is the proportion of examples for which the LSTM performs the correct classification. It is represented as the fraction of examples classified correctly. The speculated ideal model demonstrates the highest accuracy amongst all the models of the sampled candidate set which are tested. We compare in Table 2 the performance of the \emph{PDeepLearn} system with the ARMS system, both in terms of the running time of the algorithm, and the syntactic similarity of the models learnt respectively by the systems with the hand woven model used to generate the traces.
The results demonstrate that the execution time of PDeepLearn is close to that of ARMS for 700 traces.

The difference between the hand woven model and the empirically determined model is represented in the form of a \textit{reconstruction error}. This error is based on the similarity among the predicates between the empirical model and hand woven (ideal) truth model. Let $diffpre_{a_i}$ represent the syntactic difference in \textit{pre} lists of action $a_i$ in the hand woven model and the empirical model. Each time the \textit{pre} list of the ideal model presents a predicate which is not in the \textit{pre} list of the empirical model, the  count $diffpre_{a_i}$ is incremented by one. Similarly, each time the \textit{pre} list of the empirical model presents a predicate which is not in the \textit{pre} list of the ideal model, the count $diffpre_{a_i}$ is incremented by one. Similar counts are estimated for the \textit{add} and \textit{del} lists as  $diffadd_{a_i}$ and $diffdel_{a_i}$ respectively. This total count is then divided by the number of relevant constraints for that particular action $relCons_{a_i}$ to obtain the cumulative error per action. This error is summed up for every action and averaged over the number of actions of the model to obtain an average error $E$ for the entire model. The reconstruction error for the model is thus represented by:

\begin{equation}
 E = \frac{1}{n}\sum_{i=1}^{n}\frac{diffPre_{a_i}+diffAdd_{a_i}+diffDel_{a_i}}{relCons_{a_i}}
\end{equation}

The reconstruction errors are summarized in the Table 3. The empirically obtained model is identical to the hand woven one, as exhaustive generation then filtration ensures that this identical model is always part of the set and eventually narrowed down upon by the LSTM. The error $E$ of the model produced by ARMS fluctuates between 15-30 percent.



\begin{table}
\parbox{.45\linewidth}{
\centering
\begin{tabular}{|l|l|l|l|l|}
\hline
Domain    &\begin{tabular}[c]{@{}l@{}} Running time \\for ARMS\end{tabular} & \begin{tabular}[c]{@{}l@{}} Generation\\time\\for PDL\end{tabular}& \begin{tabular}[c]{@{}l@{}} Sampling\\time\\for PDL\end{tabular} & \begin{tabular}[c]{@{}l@{}} Running\\time\\for PDL\end{tabular}\\ \hline
Satellite & 29781.24 & 102.61 & 3.10 & 47857.19       \\ \hline
Mprime    & 1614.09 & 16.52 & 618.80 & 2167.07 \\ \hline
Gripper   & 36873.72 & 110.03 & 0.08 & 17253.49 \\ \hline
Depots    & 110415.61  & 153.88 & 4080.17 & 98878.02 \\ \hline
\end{tabular}
\caption{Comparison of running time for ARMS and PDeepLearn (PDL) in secs}
}
\hfill
\parbox{.47\linewidth}{
\centering
\begin{tabular}{|l|l|l|l|}
\hline
Domain    &\begin{tabular}[c]{@{}l@{}} Accuracy\\Rate (\%)\ \end{tabular}& \begin{tabular}[c]{@{}l@{}} ARMS\\Error $E$\end{tabular}\\ \hline
Satellite & 85.71 &28.88       \\ \hline
Mprime    & 75.00  & 16.66     \\ \hline
Gripper   & 100.00  & 22.22    \\ \hline
Depots    & 72.00 &24.07       \\ \hline
\end{tabular}
\caption{Accuracy Comparisons}
}
\end{table}
\section{Conclusion}
\label{conclusion}
In this paper, we have presented an approach called \emph{PDeepLearn}, which, given a set of state-action interleaved traces, uses the sequence labelling capabilities of LSTMs to learn the action model best representative of the traces. Given that the current is focused on the short term dependencies between actions, future proposed work would investigate the effect of long term dependencies in the form of constraints, and their combined implication along with the short-term dependencies on the quality of the learnt model.

\bibliography{research}

%
%
%
\bibliographystyle{splncs04}
\bibliography{mybibliography}

\end{document}